%% file: main.tex
\documentclass[lettersize,journal]{IEEEtran}
\usepackage{amsmath,amsfonts}
\usepackage{array}
\usepackage[caption=false,font=normalsize,labelfont=sf,textfont=sf]{subfig}
\usepackage{textcomp}
\usepackage{stfloats}
\usepackage{url}
\usepackage{verbatim}
\usepackage{graphicx}
\def\BibTeX{{\rm B\kern-.05em{\sc i\kern-.025em b}\kern-.08em
    T\kern-.1667em\lower.7ex\hbox{E}\kern-.125emX}}
\usepackage{balance}
\usepackage{xcolor}
\usepackage{booktabs}
\usepackage{multirow}
\usepackage{subcaption}
\usepackage{algorithm}
\usepackage{algpseudocode}
\usepackage{listings}
\usepackage{overpic}

\usepackage[pagebackref,breaklinks,colorlinks]{hyperref}

\begin{document}
\title{PAPRLE (Plug-And-Play Robotic Limb Environment): \\ A Modular Ecosystem for Robotic Limbs
}


\author {Obin Kwon$^{1*}$, Sankalp Yamsani$^{1*}$, Noboru Myers$^{1}$, Sean Taylor$^{1}$, Jooyoung Hong$^{1}$, \\
Kyungseo Park$^{2}$, Alex Alspach$^{3}$ and Joohyung Kim$^{1}$ \\
\small{ \texttt{\{obinkwon, yamsani2, noborum2, seanlt2, jh97, joohyung\}@illinois.edu} \\
 \texttt{kspark@dgist.ac.kr, alex.alspach@tri.global}}


\thanks{
$^{*}$These authors equally contributed to this work. \\
$^{1}$ Department of Electrical and Computer Engineering, University of Illinois Urbana-Champaign, Champaign, Illinois, USA. \\
$^{2}$ Department of Robotics and Mechatronics Engineering, Daegu Gyeongbuk Institute of Science and Technology (DGIST), Daegu, South Korea\\
$^{3}$ Toyota Research Institute, Los Altos, California, USA 
}
\thanks{This work was partly supported by ROBOTIS and Toyota Research Institute. Corresponding Author:  Joohyung Kim}
}


\maketitle
\thispagestyle{plain}
\pagestyle{plain}

\begin{abstract}
We introduce PAPRLE (Plug-And-Play Robotic Limb Environment), a modular ecosystem that enables flexible placement and control of robotic limbs. 
With PAPRLE, a user can change the arrangement of the robotic limbs, and control them using a variety of input devices, including puppeteers, gaming controllers, and VR-based interfaces.
This versatility supports a wide range of teleoperation scenarios and promotes adaptability to different task requirements.
To further enhance configurability, we introduce a pluggable puppeteer device that can be easily mounted and adapted to match the target robot configurations.
PAPRLE supports bilateral teleoperation through these puppeteer devices, agnostic to the type or configuration of the follower robot.
By supporting both joint-space and task-space control, the system provides real-time force feedback, improving user fidelity and physical interaction awareness.
The modular design of PAPRLE facilitates novel spatial arrangements of the limbs and enables scalable data collection, thereby advancing research in embodied AI and learning-based control. 
We validate PAPRLE in various real-world settings, demonstrating its versatility across diverse combinations of leader devices and follower robots.
The system will be released as open source, including both hardware and software components, to support broader adoption and community-driven extension. Additional resources and demonstrations are available at the project website: \url{https://uiuckimlab.github.io/paprle-pages}.
\end{abstract}


\section{Introduction}

\begin{figure}[t]
  \centering
  \includegraphics[width=\columnwidth]{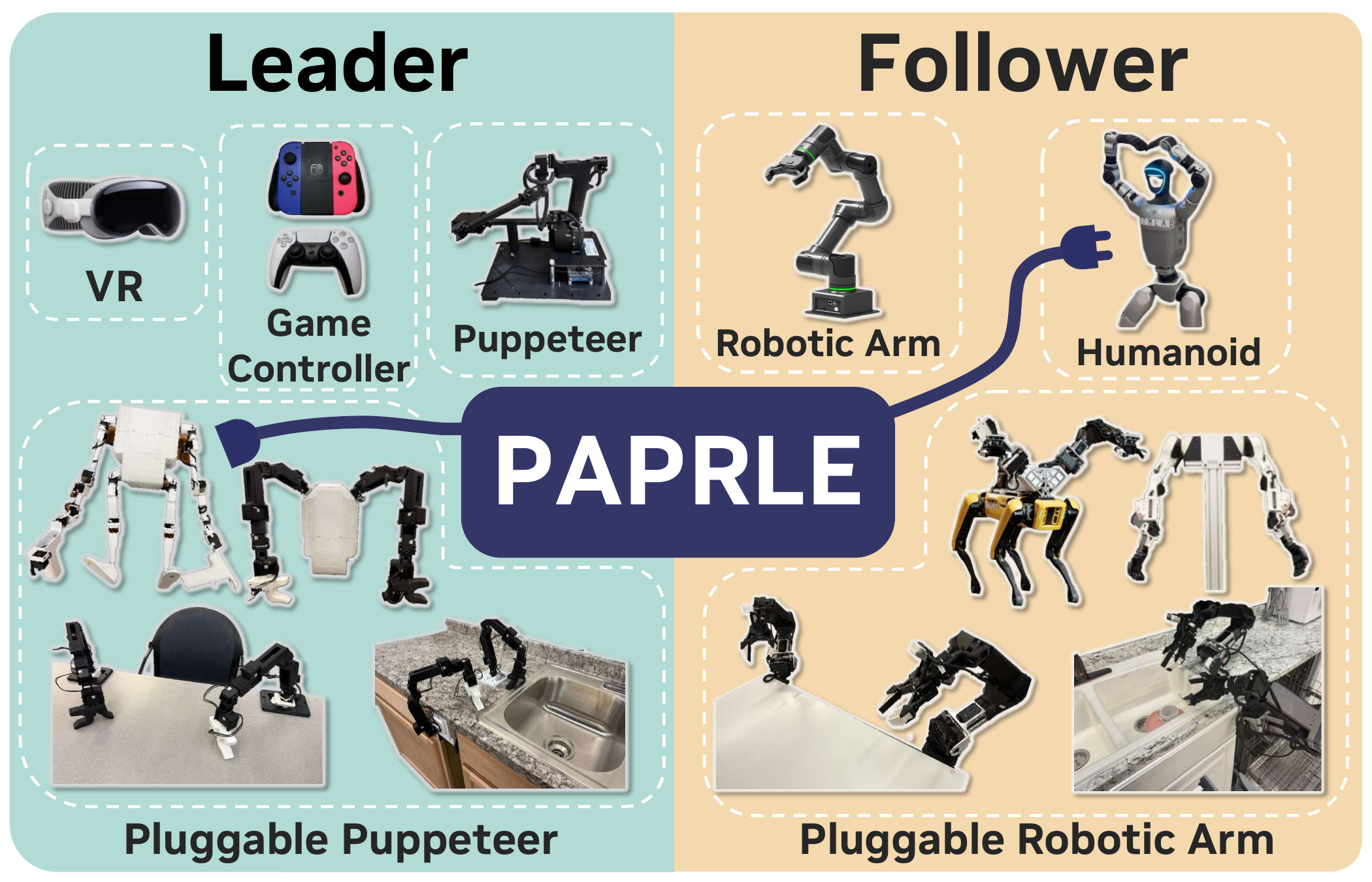}
  \caption{\textbf{Concept of PAPRLE.} Various leader devices (left) can be flexibly paired with diverse robot embodiments (right) through a plug-and-play teleoperation system that supports modular and interchangeable components.}
\label{fig:hero}
\end{figure}

In many research and deployment scenarios, the flexible arrangement of robotic limbs is often required to support a wide range of task configurations. 
For example, a table-mounted arm is well-suited for tabletop tasks, but becomes limited for more complex tasks, such as loading dishes from a sink to dishwasher, where reachability and varied interaction angles are required.
To address this limitation, the Plug-and-Play Robotic Arm System (PAPRAS) \cite{papras} was introduced, enabling modular assembly and flexible arrangement of robotic arms across diverse settings, including single-arm, or multi-arms, or even humanoid-like platforms.
The bottom-right side of Figure \ref{fig:hero} shows the various examples of PAPRAS configurations.
While PAPRAS facilitates physical modularity, a unified control method capable of seamlessly operating across these diverse configurations remains an open challenge.
Building on the modular philosophy of PAPRAS, we introduce PAPRLE (Plug-And-Play Robotic Limb Environment), a system that extends plug-and-play principles into the control domain.
We use the term ``\textit{environment}" for the proposed system because it provides a flexible and composable platform where various robot configurations (followers) and control interfaces (leaders) can be easily plugged in, mixed, and matched.
Figure \ref{fig:hero} illustrates the core concept of PAPRLE.
PAPRLE supports flexible mix-and-match pairing between leaders and followers, enabling modular and device-agnostic control across a wide range of robotic platforms.
Operators can `plug in' puppeteer devices to define a leader interface, which acts as a physical controller for the teleoperation system. 
In addition to puppeteers, other control modalities such as gaming controllers or VR devices, can also be `plugged in' as a leader. 
This ``plug-and-play'' characteristic also applied to the follower side. 
The operator can configure custom robotic limb setups using PAPARS as a follower, or they can use commercial humanoid or robotic arm platforms as a follower.

To support the variable limb configurations offered by PAPRAS, we introduce a pluggable puppeteer device, which can be freely mounted on various places. (shown in Figure \ref{fig:puppet_mount}.)
The puppeteer device is a low-cost, 3D-printed, scaled-down version of the robot, actuated by off-the-shelf motors.
Inspired by recent works \cite{gello, aloha, factr, echo}, this approach allows for direct joint-value mapping, providing users with an intuitive and physically grounded control experience. 
By attaching a mount to such puppeteer device, we made a pluggable puppeteer device.
Using this device, multiple puppeteers can be mounted as needed and also can be swapped to other puppeteers, enabling intuitive control of multi-limb robot setups.

Further, PAPRLE does not limited to direct joint mapping teleoperation using puppeteer devices. PAPRLE also supports teleoperation using a puppeteer with a different morphology from the follower robot, as well as non-puppeteer interfaces such as VR headsets or gaming controllers.
For those leader devices that do not have the same morphology with the follower robot, PAPRLE solves the inverse kinematics accordingly. In the case of puppeteer devices, even when their structure does not physically match the follower robot, the system still provides force feedback to the operator.

The composable architecture of PAPRLE facilitates scalable data collection across diverse robot-control pairings, addressing the pressing need for rich, real-world embodied interaction data.
Compared to existing teleoperation systems that often targeted specific device–robot pairings, PAPRLE aims for a broader, extensible framework.

The contributions of this paper are summarized as follows:
\begin{itemize}
    \item PAPRLE enables flexible robotic limb arrangements on both leader and follower side, ranging from tabletop arms to multi-limb humanoid systems
    \item We present a device- and robot-agnostic teleoperation system that supports diverse combinations of leader and follower devices.
    \item PAPRLE platform will be released as open source to encourage reproducibility, extensibility, and community-driven research.
\end{itemize}

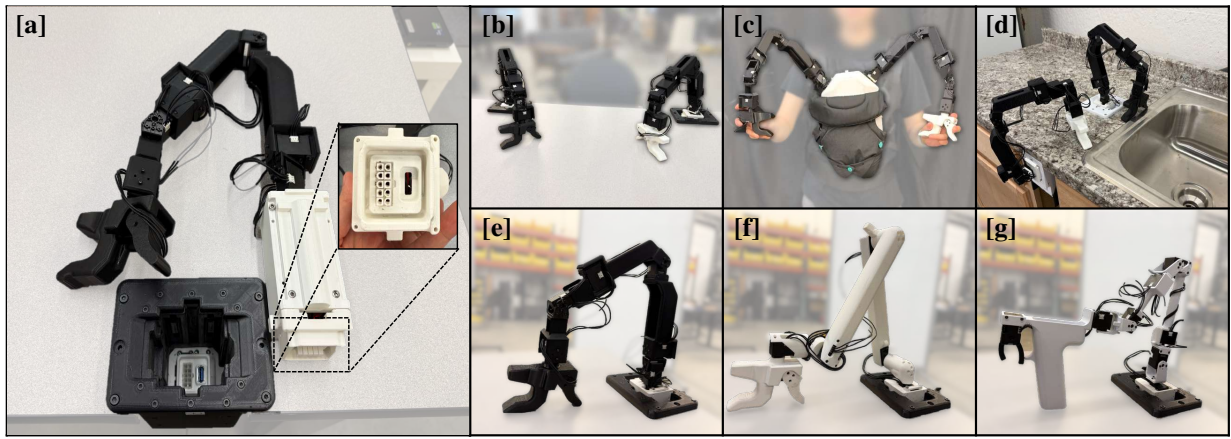
\begin{figure*}[t]
  \centering

  \label{fig:pluggable_puppet}
  \begin{overpic}[width=0.9\linewidth]{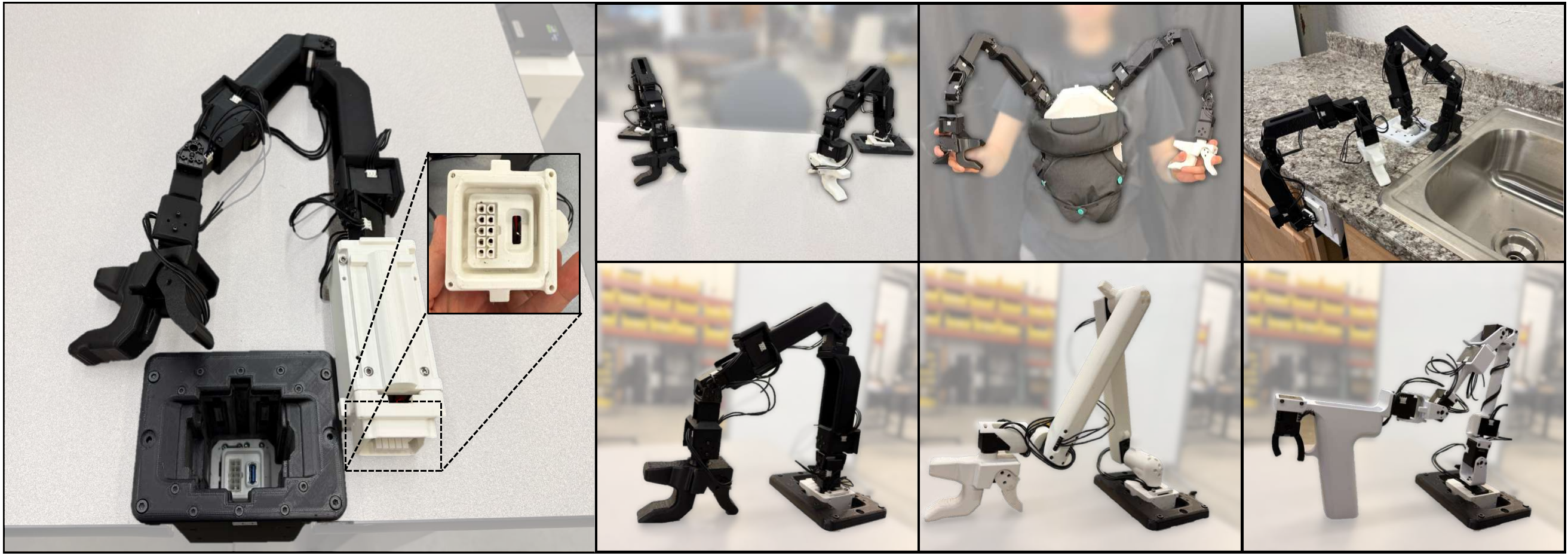}
    \refstepcounter{figure}
    \put(1,33){\refstepcounter{subfigure}\label{fig:puppet_mount}\textbf{[a]}}
    \put(39,33){\refstepcounter{subfigure}\label{fig:puppet_table}\textbf{[b]}}
    \put(59.5,33){\refstepcounter{subfigure}\label{fig:puppet_orthrus}\textbf{[c]}}
    \put(80,33){\refstepcounter{subfigure}\label{fig:puppet_kitchen}\textbf{[d]}}
    \put(39,16.5){\refstepcounter{subfigure}\label{fig:puppet_7dof}\textbf{[e]}}
    \put(59.5,16.5){\refstepcounter{subfigure}\label{fig:puppet_ur5}\textbf{[f]}}
    \put(80,16.5){\refstepcounter{subfigure}\label{fig:puppet_baby}\textbf{[g]}}

  \end{overpic}
  \addtocounter{figure}{-1}
  \caption{\textbf{Pluggable Puppeteers} (a) The proposed puppeteer device and mounting interface. (b–d) Examples of the device mounted in diverse scenarios. (e–g) The same mounting base accommodates a variety of puppeteer devices, which are scaled replicas of different arms such as PAPRAS, UR5, and Unitree G1's arm, respectively.}
\end{figure*}

\section{Related Work}

A growing movement in robotics research has focused on teleoperation as a pathway to large-scale data collection for learning-based robot policy.
Systems such as ALOHA~\cite{aloha}, Gello \cite{gello} enable intuitive teleoperation through puppeteer devices by replicating the kinematics of the target robot.
On the other hand, BunnyVisionPro~\cite{bunny-visionpro} or OpenTeleVision~\cite{open-tv} proposed a VR-based pipeline for capturing human wrist pose and dextrous hand motion and translating it into robot control signals.
In addition, FACTR \cite{factr} and Echo \cite{echo} further enhance joint-space teleoperation with bilateral force feedback, enabling high-quality demonstrations for contact-rich tasks. 
While these approaches have shown compelling results, they are often designed for fixed device–robot pairings, limiting their adaptability to diverse teleoperation scenarios.

\vspace{-0.1cm}
Focusing on broader applicability, several works have proposed modular and cross-platform teleoperation architectures that allow flexible pairing of input devices (“leader”) with a range of robotic platforms (“follower”). 
ACE \cite{ACE} introduces a portable exoskeleton with vision-based tracking, supporting cross-robot teleoperation across hands, arms, and quadrupeds.
TeleMoMa \cite{telemoma} allows users to control mobile manipulators using various input devices such as RGB-D cameras or gamepads by abstracting user actions into a shared control space. 
MoMa-Teleop \cite{wholebodyteleop} takes a hybrid approach where the user controls the end-effector while the robot autonomously adjusts its base for extended reach. 
These systems demonstrate increased input modularity and cross-platform compatibility, but typically target a narrow set of follower configurations, often using commercial robots or predefined setups. 

In contrast, our system supports arbitrary limb configurations on both the leader and follower sides, enabling plug-and-play reconfiguration across diverse embodiments. 
The key is the plug-and-play abstraction: not only can users swap between input devices (e.g., 3D-printed puppeteers, game controllers, VR), but the proposed puppeteer leader device itself is also modular and pluggable. 
This flexibility propagates to the follower side, where a wide range of robots from tabletop scenarios to humanoids, can be teleoperated using the same unified interface. 
Furthermore, we extend beyond existing work by providing force feedback even in heterogeneous scenarios, where the puppeteer leader and follower do not share the same structure or degrees of freedom.

\section{Pluggable Puppeteers}
In this section, we introduce a pluggable puppeteer leader device, designed for PAPRLE.
Similar to Gello \cite{gello}, this device can be designed as a scaled replica of a specific follower robot, fabricated using 3D-printed components. 
Equipped with low-cost motors, it can read joint states in real time and relay them to control the follower robot.
The device has a mount, which can be easily mounted and removed in various places, enabling flexible deployment across various setups.
Figure \ref{fig:puppet_mount} illustrates the proposed puppeteer device alongside its mounting mechanism. 
The mount design is similar to the one from PAPRAS, which allows the same mounts to be used interchangeably between PAPRAS and puppeteer devices, enabling more diverse and reconfigurable setups.
TTL communication and power are routed through the mount to the leader device, which is secured with a simple retaining bar. This allows for quick swapping of leader devices using a common mounting base.
For more detailed information on the mounting mechanism, please refer to  \cite{papras}, \cite{baby}.

Using this pluggable design, multiple types of puppeteer device can be interchanged and arranged in different configurations.
Figure \ref{fig:puppet_table}, \ref{fig:puppet_kitchen}, and \ref{fig:puppet_orthrus} show example configurations using the same set of puppeteer devices.
The reconfiguration process is simple; users can simply unplug a device from one mount and plug it into another.
The same mount can also accommodate entirely different types of arm.
Figure \ref{fig:puppet_7dof}, \ref{fig:puppet_ur5}, and \ref{fig:puppet_baby} show examples of using different puppeteer devices for the same mount.
This plug-and-play functionality of leader devices enables users to swap between different puppeteer devices or configurations, supporting a wide range of manipulation scenarios with minimal setup time.

\begin{figure*}[t]
  \centering
  \includegraphics[width=0.9\linewidth]{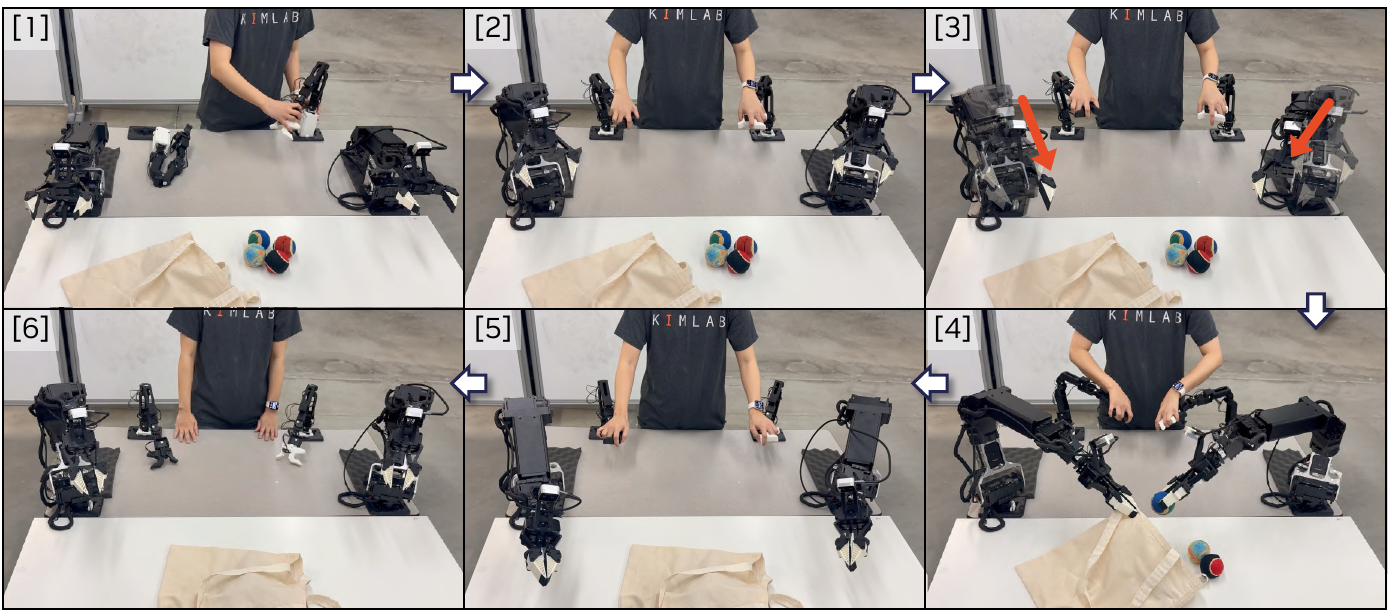}
  \caption{\textbf{Example of Teleoperation Process using Pluggable Puppeteers.}}
\label{fig:teleop}
\end{figure*}

\begin{figure*}[t]
  \centering
  \includegraphics[width=\linewidth]{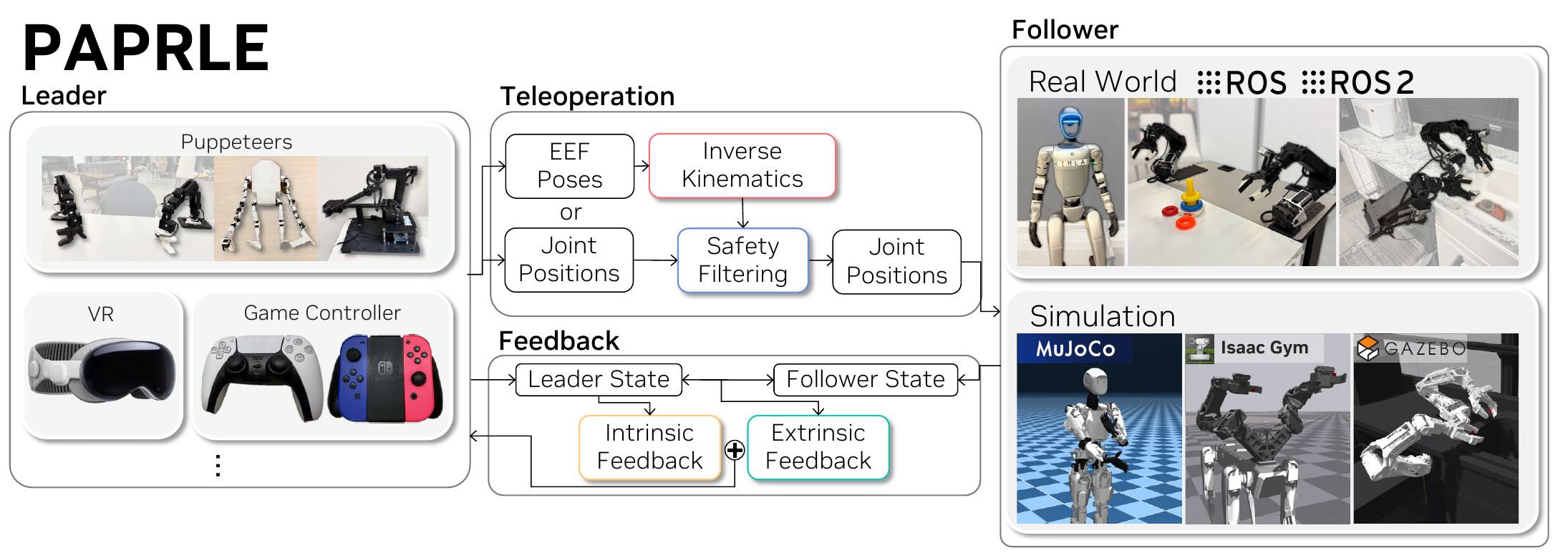}
  \caption{\textbf{Overview of PAPRLE Teleopeartion System.}}
\label{fig:overview}
\end{figure*}

\section{PAPRLE Teleoperation System}
%

\subsection{Teleoperation Session Workflow} \label{sec:teleop_workflow}

\begin{algorithm}[t]
\caption{Teleoperation Session}
\label{alg:teleop}
\include{tables/teleop_pseudo}
\vspace{-0.5cm}
\end{algorithm}

The overall procedure of a teleoperation session is illustrated in the pseudo code shown in Algorithm \ref{alg:teleop}, and a step-by-step illustration is shown in Figure \ref{fig:teleop}.
PAPRLE consists of four high-level modules: (1) \texttt{leader}, (2) \texttt{teleoperation}, (3) \texttt{follower} and (4) \texttt{feedback}.

To initiate a teleoperation session, the user must specify the leader device, the follower robot, and the target environment.
Based on these inputs, the system selects the appropriate configuration files and uses them to set up each module.
For example, as in Figure \ref{fig:teleop}, if a user wants to control PAPRAS robots using puppeteer devices in the ROS2 environment, the selected configurations would be:
(1) \texttt{$\text{cfg}_\text{leader}$} for the puppeteer devices,
(2) \texttt{$\text{cfg}_\text{follower}$} for the PAPRAS robots, and
(3) \texttt{$\text{cfg}_\text{env}$} for the ROS2 environment.
To switch from ROS2 environment to simulation setup, the user can simply change $\text{cfg}_\text{env}$ to the corresponding simulator configuration file, while using the same leader and follower configuration files.
To setup a new follower robot, an URDF file is needed and the user just needs to write a corresponding configuration file.

At the start of each session, the follower robot is initialized with a predefined base pose. The robot moves to this pose and remains idle until all other modules are fully initialized.
Once all module initializations are complete, the system awaits a start signal from the operator, corresponding to step 2 in Figure \ref{fig:teleop} and line 6 in Algorithm \ref{alg:teleop}.
In step 2 of Figure \ref{fig:teleop}, the operator initiates the process by grasping both grippers of the puppeteer devices as the start signal.
Upon leader module detects the start signal, the follower robot slowly moves to the posture mirroring the operator's pose, aligning with step 3 in Figure \ref{fig:teleop} and line 9 in Algorithm \ref{alg:teleop}.

The teleoperation session is structured as a simple for-loop, following a format similar to the gym environment\footnote{\url{https://github.com/openai/gym}}, consisting of repeated observation and action selection. This design is intentionally adopted to facilitate future integration with AI agents.
During each iteration, the leader module provides commands, which are processed by the teleoperation module.
Teleoperation module translates the commands from the leader into joint-space actions, and the resulting command is sent to the follower module to control the robot.
Concurrently, the feedback module runs in the background, relaying feedback to the leader module.
Figure \ref{fig:overview} shows the illustrated diagram of each module. 
Details for each module will be discussed in the following subsections.

Each leader device features its own end signal, configurable via separate classes. The teleoperation loop simply monitors a Boolean variable from leader module to detect when a leader device signals the end of a teleoperation session, triggering a reset of the follower robot.
In Figure \ref{fig:teleop} step 5, the operator sends end signals to the leader module by grasping the grippers in a specific pose.
After reset, the system will wait for the next start signal, allowing for repeated teleoperation sessions.

\subsection{Leader Module} \label{subsec:leader_module}
The leader module serves as the interface between each leader device and the PAPRLE system.
This module translates input signals from the leader device into appropriate command formats for the teleoperation module. 
Depending on the configuration of the paired follower robot, the leader module outputs either (1) joint positions or (2) end-effector poses represented as $\mathrm{SE(3)}$ transformations.
When the puppeteer has the same joint configuration as the follower robot, the leader module outputs joint positions as commands. 
However, one-to-one joint mapping is not feasible when a different type of puppeteer device is used as the leader, or when the device does not have joint positions such as gaming controller or VR device.
In these cases, leader module outputs end-effector poses, and the following teleoperation module performs inverse kinematics (IK) to compute joint-level commands for the follower robot.
The end-effector pose $\mathbf{T}$ at time step $t$ is defined as a homogeneous transformation matrix:
\begin{equation}
\mathbf{T}_{t} =
\begin{bmatrix}
\mathbf{R} & \mathbf{t} \\
\mathbf{0} & 1
\end{bmatrix},
\end{equation}
where $\mathbf{R} \in SO(3)$ is the rotation matrix representing orientation, and $\mathbf{t} \in \mathbb{R}^3$ is the translation vector representing position.
At the start of a teleoperation session ($t=0$), the leader module stores the initial end-effector pose of the leader device, denoted as $\mathbf{T}^{\text{leader}}_0 \in SE(3)$, while the teleoperation module stores the initial end-effector pose of the follower robot as $\mathbf{T}^{\text{follower}}_0$.
If the leader device is a puppeteer, the end-effector pose is computed via forward kinematics using the URDF model. 
For hand-held devices, the initial pose can be set to the identity matrix $\mathbf{I}$. 
For VR-based leaders, the human wrist pose is used as the end-effector pose.
During the teleoperation session, the leader module continuously computes its current end-effector pose, denoted as $\mathbf{T}^{\text{leader}}_{0 \rightarrow t} \in SE(3)$.  
The control command is defined as the pose delta from the initial pose:
\begin{equation}
\begin{split}
\mathbf{T}^{\text{leader}}_{0 \rightarrow t} &= \left( \mathbf{T}^{\text{leader}}_0 \right)^{-1} \mathbf{T}^{\text{leader}}_t \\
&=
\begin{bmatrix}
\mathbf{R}^{\text{leader}}_{0 \rightarrow t} & \mathbf{t}^{\text{leader}}_{0 \rightarrow t} \\
\mathbf{0} & 1
\end{bmatrix},
\end{split}
\end{equation}
where $\mathbf{R}^{\text{leader}}_{0 \rightarrow t}$ denotes the relative rotation and $\mathbf{t}^{\text{leader}}_{0 \rightarrow t}$ denotes translation from time $0$ to $t$.

To account for the differences in scale between the leader and follower devices, a scalar factor $s \in \mathbb{R}$ is applied to the translational component of the pose delta. The scaled pose delta is then defined as:
\begin{equation}
\widetilde{\mathbf{T}}^{\text{leader}}_t =
\begin{bmatrix}
\mathbf{R}^{\text{leader}}_{0 \rightarrow t} & s \cdot \mathbf{t}^{\text{leader}}_{0 \rightarrow t} \\
\mathbf{0} & 1
\end{bmatrix}.
\end{equation}
The scaled delta $\widetilde{\mathbf{T}}^{\text{leader}}_t$ is then forwarded to the teleoperation module and applied to the follower’s initial pose to compute the follower’s target pose:
\begin{equation}
\mathbf{T}^{\text{follower}}_{t,cmd} = \mathbf{T}^{\text{follower}}_0 \cdot \widetilde{\mathbf{T}}^{\text{leader}}_t.
\label{eq:follower_target_pose}
\end{equation}
By adopting this delta-based end-effector command, our teleoperation framework enables seamless integration of diverse leader devices.

\begin{table}[t]

\caption{\textbf{Supported Leader Devices.}}
\centering
\input{tables/supported_devices}
\label{tab:devices}
\end{table}

The list of supported leader devices is shown in Table \ref{tab:devices}, categorized by device type, supported command format, number of controllable limbs.
\begin{itemize}
    \item \textbf{Puppeteer devices} replicate a scaled-down structure of a specific target follower robot. The list is the devices we have tested, including UR5 from Universal Robotics\footnote{\url{https://www.universal-robots.com/products/ur5e/}}, and Open Manipulator Y model (OM-Y) from ROBOTIS\footnote{\url{https://www.dynamixel.com/omy.php}}. Users can also design and use their own puppeteer devices.
    \item \textbf{VisionPro},  a VR headset from Apple, provides estimated human wrist poses, enabling end-effector pose control of up to two limbs. This device is particularly useful when a lightweight and portable teleoperation setup is preferred.
    \item \textbf{Gaming Controller} such as Joycon from Nintendo Switch\footnote{\url{https://www.nintendo.com/store/hardware/joy-con-and-controllers/}} and DualSense from PlayStation 5\footnote{\url{https://www.playstation.com/accessories/dualsense-wireless-controller/}}.
    \item \textbf{Offline trajectory} can also serve as leader inputs, typically in the form of pre-recorded datasets streamed into the system.

\end{itemize}
Importantly, our system does not assume a fixed number of limbs.
PAPRLE supports multi-limb control within a single teleoperation session, enabling flexible configurations. 
However, the number of limbs that can be controlled simultaneously may be constrained by the capabilities of the leader device.
For example, the gaming controller can handle only one limb per session.
In case of pluggable puppeteer devices, designed to be mountable in the same plug-and-play manner as PAPRAS arms, the leader configuration can be physically matched to that of the follower robot.
For example, two puppeteers can be mounted for a dual-arm follower robot, while four puppeteers can enable full-limb teleoperation of a humanoid robot.
Figure \ref{fig:follower_g1_full} shows an example of such multi-limb configuration.
Moreover, the system is easily extensible: a new type of leader device can be integrated by implementing a simple class that periodically publishes commands in the appropriate format, allowing rapid support for novel hardware interfaces.

\subsection{Teleoperation Module} \label{subsec:teleoperation_module}

The teleoperation module receives commands from the leader device and translate them into a control signal for the follower. 
This process consists of two main steps.
\paragraph{Command Interpretation}
If the command is given as a delta end-effector pose $\widetilde{\mathbf{T}}_t^{\text{leader}}$, the module first computes the resulting target pose for the follower robot’s end-effector $\mathbf{T}^{\text{follower}}_{t,\text{cmd}}$ using Equation \ref{eq:follower_target_pose}. 
To translate this command into joint positions of the follower robot, the module solves an inverse kinematics (IK) problem using a Jacobian-based iterative algorithm. 
Given the forward kinematics function $f$, the IK solver aims to find the joint positions $\mathbf{q}^*$ such that the resulting end-effector pose $\mathbf{f}(\mathbf{q})$ closely matches the desired pose $\mathbf{x}_d$. 
Starting from the initial guess $\mathbf{q}_0$, the solver iteratively updates the joint positions by computing the pose error $\mathbf{e}$, and mapping it into the joint space using the Jacobian $\mathbf{J}(\mathbf{q})$ as: \begin{equation} \Delta \mathbf{q} = -\mathbf{J}(\mathbf{q})^{\dagger} \mathbf{e}, \end{equation} where $\mathbf{J}(\mathbf{q})^{\dagger}$ denotes the damped pseudo-inverse of the Jacobian.

\begin{figure}
    \centering
    \includegraphics[width=\columnwidth]{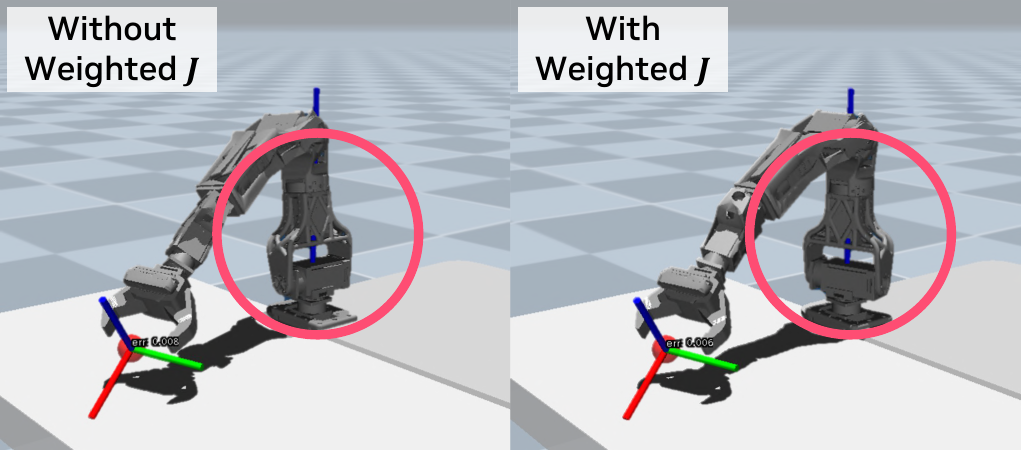}
    \caption{\textbf{Examples of Differences in IK Results.}}
    \label{fig:constrained_ik}
\end{figure}

To make this process more suitable for teleoperation, we made several design choices. 
Since this is a real-time teleoperation scenario, we can assume that the end-effector does not move significantly between consecutive commands. 
Thus, we set the initial guess $\mathbf{q}_0$ as the current joint position of the follower robot for faster convergence. 
Additionally, when the leader and follower have different kinematic structures, the end-effector movement from the leader may not directly map to a natural or desirable motion for the follower. 
Even if an IK solution is found, it can sometimes lead to undesirable postures.
Figure \ref{fig:constrained_ik} illustrates such an example: both robots reach the same end-effector pose, but with different postures. 
In the first example, the first joint is excessively tilted, resulting in unnecessary large base movements. 
Due to the structure of the Jacobian and the robot kinematics, solving IK without weighting tends to result in proximal (earlier) joints contributing more significantly to the motion.
To mitigate this, we introduce a weighting matrix $\mathbf{P}$ into the Jacobian to favor the movement of specific joints: \begin{equation} \Delta \mathbf{q} = -\left(\mathbf{P}\mathbf{J}(\mathbf{q})\right)^{\dagger} \mathbf{P} \mathbf{e}, \end{equation} where $\mathbf{P}$ is a diagonal matrix that encodes user-defined priorities for different degrees of freedom. 
For example, in Figure \ref{fig:constrained_ik}, the base joint (Joint 1) is down-weighted by setting the corresponding diagonal entry in $\mathbf{P}$ to 0.5, thus discouraging excessive movement of the base and favoring distal joints.

Using this IK algorithm, the teleoperation module translates the leader command into the joint positions of the follower robot.
%
%
This design allows the module to accommodate different types of leader devices, depending on whether they operate in task space or joint space.
If the command is already provided in joint space, this step is skipped, and the follower directly tracks the commanded joint positions.

\paragraph{Safety Filtering}
After interpretation into joint positions, the module performs safety checks, including collision detection and joint limits.
It also constrains the maximum velocity of each joint to ensure safety. 
These safety parameters are specified in \texttt{$\text{cfg}_\text{follower}$}, which are tunable by operator. 
Outputs from this filtering becomes the control signal for follower robots.

\subsection{Follower Module} \label{subsec:follower_module}
The follower module receives joint position commands and sends them to the follower robot. 
For simulation, simulator-specific functions are implemented to set the joint positions of the robot.
The supported environments and robot hardware are described in Section \ref{sec:supported_envs}. 
For hardware control, we leverage the Robot Operating System (ROS) to simplify integration and implementation. 
Since ROS provides a standardized interface for communication between software and hardware, we can easily integrate a wide range of robots for PAPRLE, utilizing existing drivers and tools. 
This reduces the need for low-level programming and enables rapid development across different robot platforms.
%


\subsection{Feedback Module} \label{subsec:feedback_module}
Feedback module is not included in the teleoperation loop, rather this module runs as a separate thread asynchronously, to ensure fast and smooth feedback signals to the leader device.
This module gathers information from the leader module and the follower module, and calculates corresponding feedback to the leader device.
PAPRLE currently provides feedback tailored for puppeteer devices, with plans to potentially integrate diverse feedback types for other leader devices in the future.

\begin{figure}
\includegraphics[width=\columnwidth]{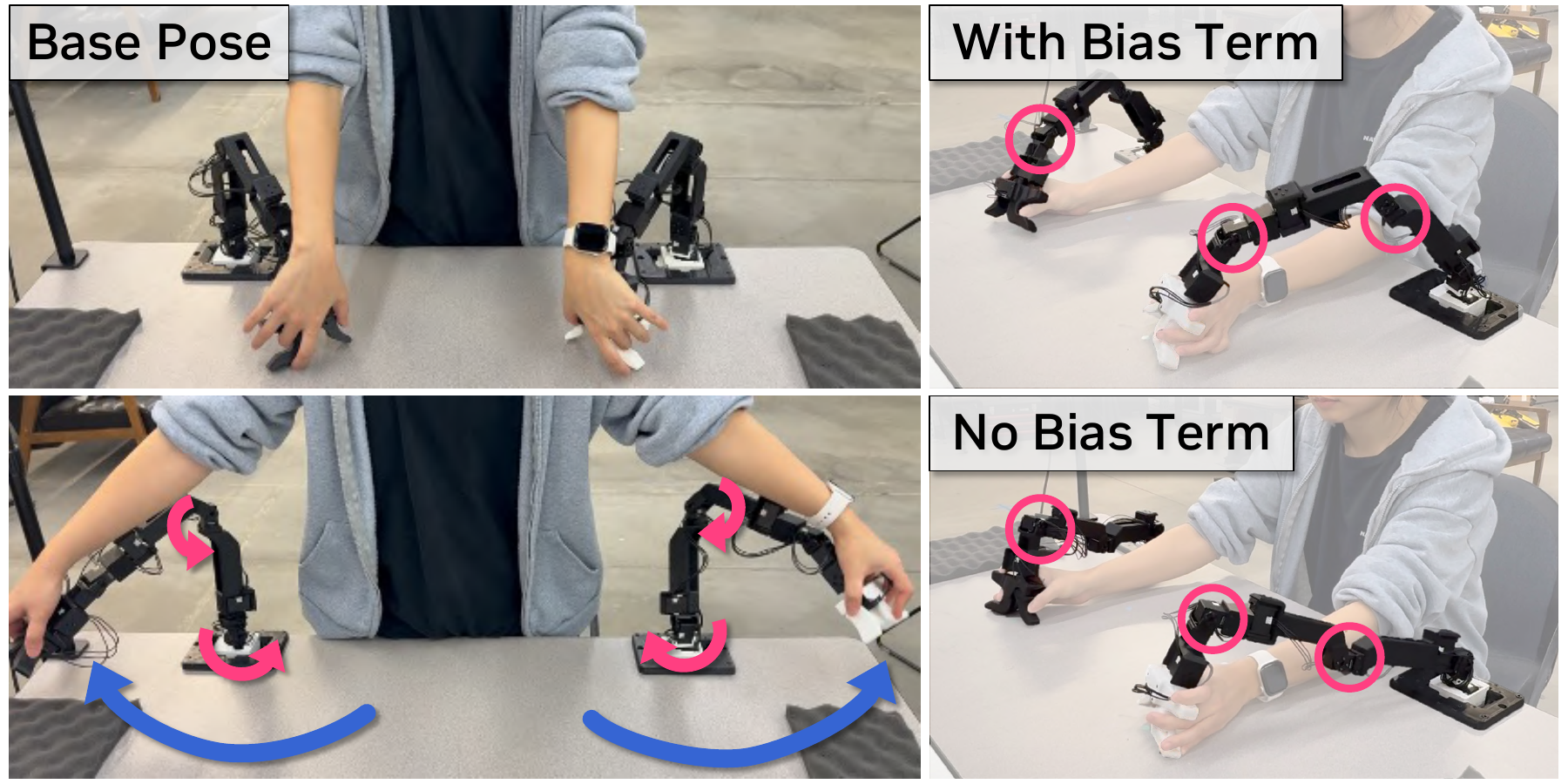}
\caption{\textbf{Example of Intrinsic Feedback.}} 
\label{fig:intrinsic_feedback}
\vspace{-0.5cm}
\end{figure}
We categorized the feedback for puppeteers into two types, (1) intrinsic feedback and (2) extrinsic feedback. 
Intrinsic feedback is solely based on the leader model, especially for puppeteer devices.
Extrinsic feedback comes from the difference between the follower and the leader.
The main form of intrinsic feedback is a biasing mechanism for puppeteer devices.
Each leader puppeteer has a predefined base pose according to its configuration, and the biasing force term is computed as
\begin{equation}
\mathbf{\tau}_\text{bias} = \mathbf{q}_\text{leader}(t) - \mathbf{q}^{\text{base}}_\text{leader}.
\end{equation}
This feedback applies a compensatory force to the puppeteer, reducing the physical effort required by the user during teleoperation.
Figure \ref{fig:intrinsic_feedback} illustrates how this compensation force works.
When the user’s motion causes the elbow of the leader to bend in an unnatural direction, the system generates a corrective torque that gently guides it back toward the base pos. 
This bias term helps the operator to maintain a natural posture of the leader device. 
Additionally, it provides an implicit hint about the workspace, as the compensatory force helps guide the user back toward a comfortable and effective control region.

\paragraph{Extrinsic Feedback}
Extrinsic feedback is based on the relationship between the leader device and the follower robot.
The feedback module continuously monitors the follower's actual pose and provides corrective feedback to the leader device when the follower fails to track the commanded motion accurately.
For example, if the follower is not following because of the velocity limits or is constrained by internal collision checking within the teleoperation module, a discrepancy may happen between the target pose and the actual executed pose.
When the command type is joint position, which means that the leader and the follower have the same joint configuration, the feedback can be simply computed as the direct difference in joint positions:
\begin{equation}
\mathbf{\tau}_\text{tracking} = \mathbf{q}_\text{leader}(t) - \mathbf{q}_\text{follower}(t) .
\end{equation}
This signal can then be used to guide the leader device toward the actual configuration of the follower, effectively informing the user of the mismatch.
This feedback naturally conveys interaction forces in the gripper, as contact with an object usually results in a deviation between the intended and actual pose of the gripper of the follower.

Even when the leader and follower robots differ in joint configuration, the feedback module can still generate force feedback using task-space representations. 
Specifically, we compute their relative difference and it is then mapped into the Lie algebra $\mathfrak{se}(3)$ using the logarithmic map, yielding a 6D twist vector that captures both translational and rotational error:
\begin{equation}
\Delta \mathbf{x}_{\text{eef}} = \log\left( {\mathbf{T}_t^{\text{follower}}}^{-1} \mathbf{T}^{\text{follower}}_{t,\text{cmd}} \right).
\end{equation}
To reflect this error on the leader device, the task-space delta is projected into the leader’s joint space using its Jacobian:
\begin{equation} \label{eq:tracking_feedback_delta_q}
\Delta \mathbf{q}_{\text{leader}} = \mathbf{J}_{\text{leader}}^{\dagger} \, \Delta \mathbf{x}_{\text{eef}},
\end{equation}
where $\mathbf{J}_{\text{leader}}^{\dagger}$ denotes the pseudo-inverse of the Jacobian. 
This joint-space correction can be used to generate torque-based feedback by applying a proportional gain $K_p$:
\begin{equation}
\mathbf{\tau}_{\text{tracking}} = -K_p \, \Delta \mathbf{q}_{\text{leader}}.
\end{equation}
This formulation enables our system to provide force feedback on the leader side even when the follower has a different morphology.
%

\begin{figure}[t]
    \centering
    \includegraphics[width=0.9\columnwidth]{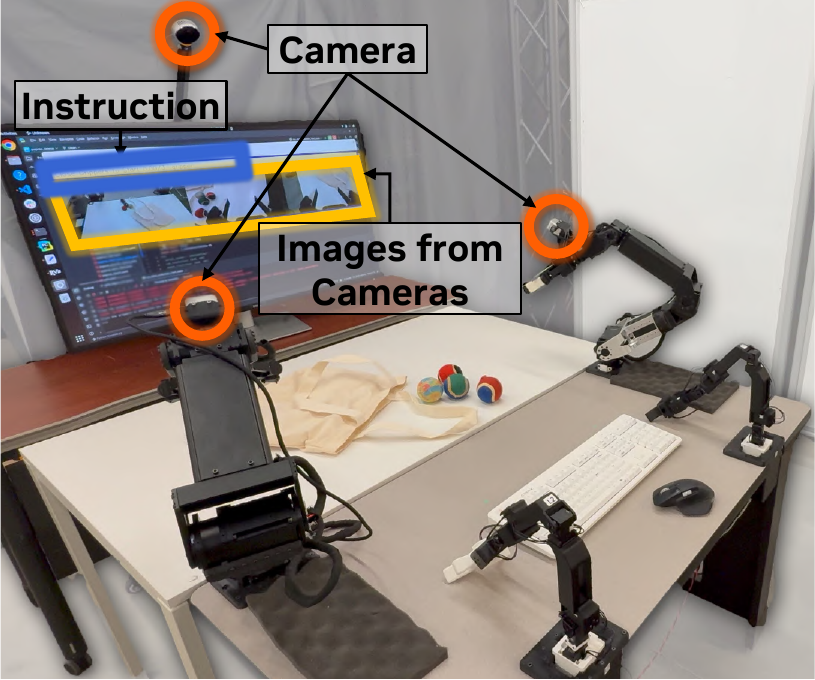}
    \caption{\textbf{Example of Data Collection Setup using PAPRLE}.}
    \label{fig:table_setup}
\end{figure}

\begin{figure*}[t!]
        \centering
    \begin{overpic}[width=\textwidth]{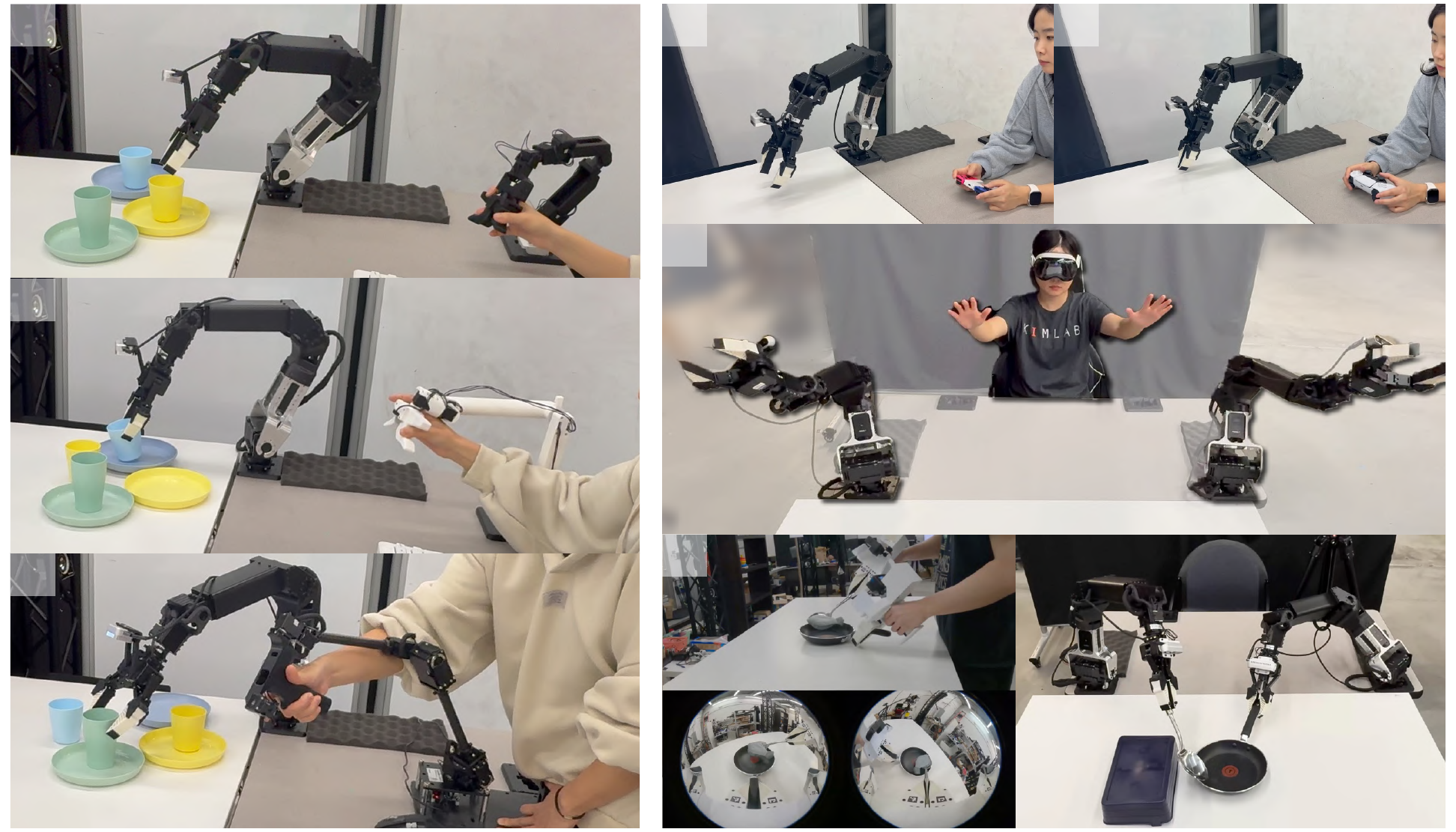}
        \refstepcounter{figure}
        \put(1.1,55.0){\refstepcounter{subfigure}\label{fig:leader_7dof}\textbf{[a]}}
        \put(1.1,36.0){\refstepcounter{subfigure}\label{fig:leader_ur5}\textbf{[b]}}
        \put(1.1,17.25){\refstepcounter{subfigure}\label{fig:leader_omy}\textbf{[c]}}
        \put(45.7,55.0){\refstepcounter{subfigure}\label{fig:leader_joycon}\textbf{[d]}}
        \put(72.8,55.0){\refstepcounter{subfigure}\label{fig:leader_dualsense}\textbf{[e]}}
        \put(45.8,39.9){\refstepcounter{subfigure}\label{fig:leader_visionpro}\textbf{[f]}}
        \put(45.8,18.55){\refstepcounter{subfigure}\label{fig:leader_umi}\textbf{[g]}}
    \end{overpic}
    \addtocounter{figure}{-1}
    \caption{\textbf{Examples of different devices as leader for controlling PAPRAS arm.} \textbf{[a-c]} Puppeteer devices designed for PAPRAS, UR5, and Open Manipulator Y model, respectively. \textbf{[d-e]} Gaming controllers (Joycon and Dualsense Controller), \textbf{[f]} VR device (Apple VisionPro), \textbf{[g] } Pre-collected dataset using UMI \cite{umi}.}
    \label{fig:diverse_leaders}
\end{figure*}

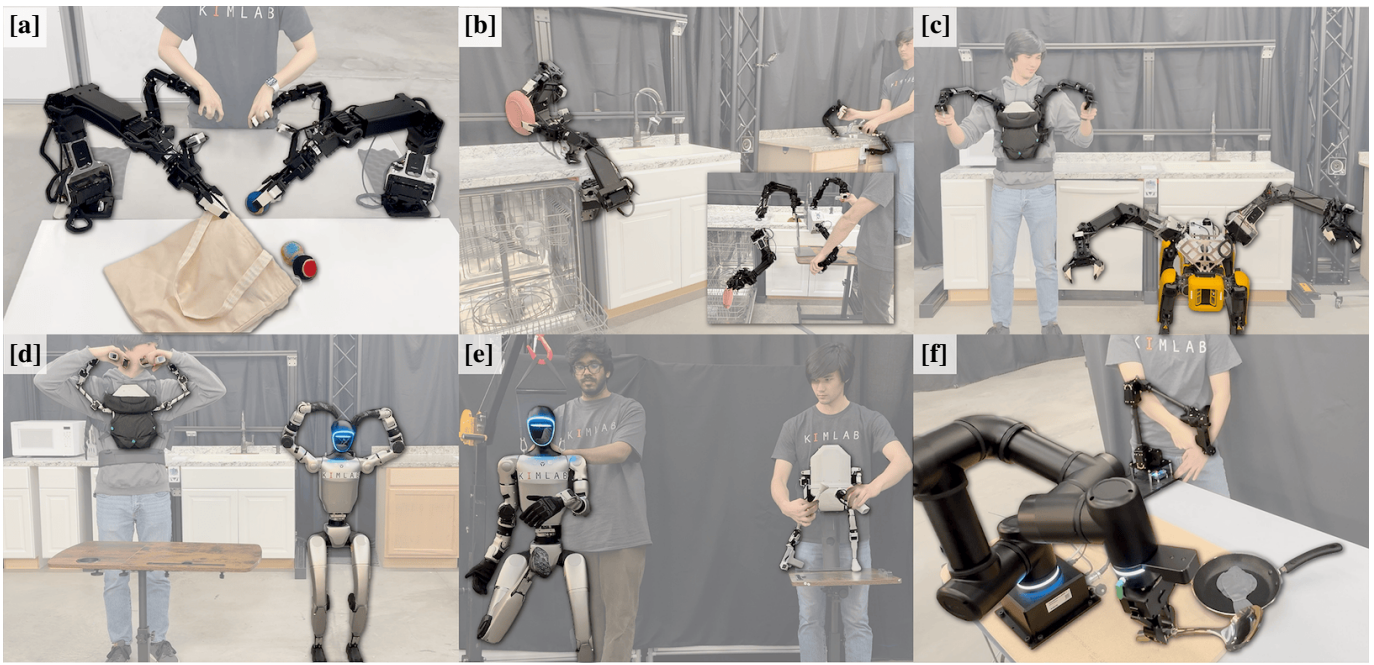
\begin{figure*}[t!]
    
    \centering
    \begin{overpic}[width=\textwidth]{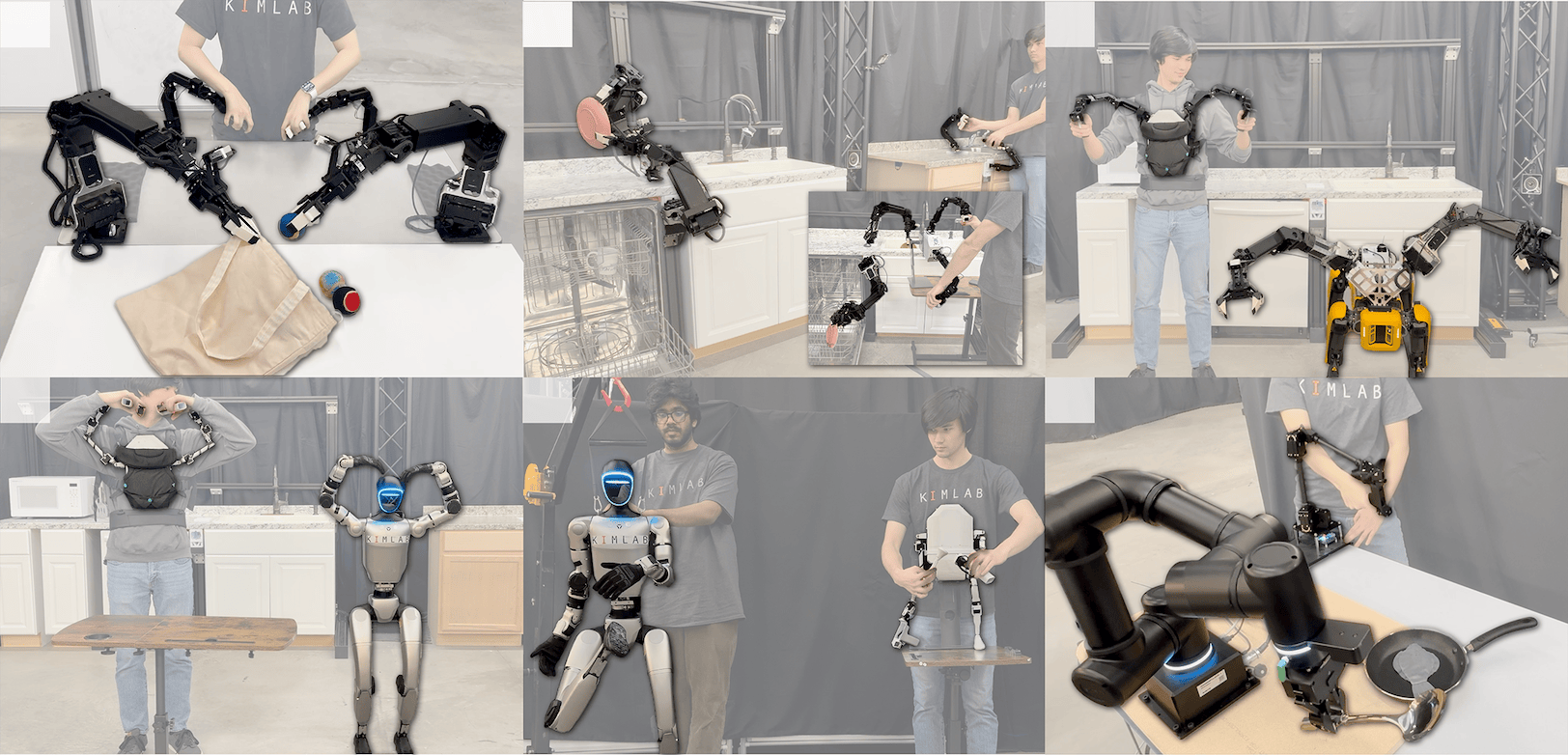}
        \refstepcounter{figure}
        \put(0.4,46.25){\refstepcounter{subfigure}\label{fig:follower_table}\textbf{[a]}}
        \put(33.75,46.25){\refstepcounter{subfigure}\label{fig:follower_kitchen}\textbf{[b]}}
        \put(67.25,46.25){\refstepcounter{subfigure}\label{fig:follower_orthrus}\textbf{[c]}}
        \put(0.4,22.1){\refstepcounter{subfigure}\label{fig:follower_g1_upper}\textbf{[d]}}
        \put(33.75,22.1){\refstepcounter{subfigure}\label{fig:follower_g1_full}\textbf{[e]}}
        \put(67.25,22.1){\refstepcounter{subfigure}\label{fig:follower_omy}\textbf{[f]}}
    \end{overpic}
    \addtocounter{figure}{-1}
    \caption{\textbf{Examples of diverse follower configurations controlled by PAPRLE.} \textbf{[a-c]} Diverse configurations using two PAPRAS arms \textbf{[d-e]} Humanoid robot (Unitree G1): dual-arm control and full-limb control \textbf{[f]} Commercial Robotic Arm (ROBOTIS Open Manipulator Y)}
    \label{fig:diverse_followers}

\end{figure*}

\section{Supported Follower Robots, Environments} \label{sec:supported_envs}

The proposed PAPRLE system enables users to operate a wide range of robots across diverse environments.
For hardware, we support both ROS1 \cite{ROS1} and ROS2 \cite{ROS2} which facilitates the seamless integration of new robots. For simulation, our system supports MuJoCo \cite{mujoco}, Isaac Gym \cite{isaacgym}, and Gazebo \cite{gazebo} (for ROS)), providing flexibility for rapid prototyping and experimentation.
We validated the system on various configurations of PAPRAS as well as the Unitree G1 robot and Open Manipulator Y from ROBOTIS. 
As part of the PAPRLE development, we also created a 7-DOF variant of PAPRAS, adding an additional joint to the original design. All PAPRAS models, including 3D-printable components and hardware interfaces, will be released as open-source.

\section{Demonstration}

In this section, we show the use cases of the proposed PAPRLE, with each example illustrated in the accompanying video.

\subsection{Example Teleoperation Setup for Data Collection}
First, we can use the proposed PAPRLE for data collection in robot learning. 
The example setup is shown in Figure \ref{fig:table_setup}. 
We mounted two PAPRAS on a table as a follower, and also mounted the puppeteer devices on the same table accordingly.
Each PAPRAS arm supports one camera to be connected through USB, and we can also connect external cameras to the system.
When a teleoperation session is launched, PAPRLE shows the real-time images from the connected cameras, collision status, and instruction for the operator. 
For example, in the Figure \ref{fig:table_setup}, the pop-up window tells the operator to `close the grippers to start teleoperation.'
PAPRLE automatically records data with timestamps in the background at each step.
This collected dataset can be processed and utilized for training a manipulation policy.
Additionally, users have the flexibility to specify which data to collect.
%

\subsection{Diverse Leaders}
With PAPRLE, we can control the same robot with diverse types of leader devices. 
Figure \ref{fig:diverse_leaders} shows an example of controlling PAPRAS arms with different types of leaders.
In Figure \ref{fig:leader_7dof}, the operator is using the puppeteer which has the same joint configuration as the PAPRAS.
In this case, the joint positions of the leader are directly forwarded to the follower robot.
In Figure \ref{fig:leader_ur5}, the operator is using the UR5 puppeteer, which is adapted from \cite{gello}, and Figure \ref{fig:leader_omy} is the off-the-shelf leader device from ROBOTIS, which has the same configuration as Open Manipulator-Y robot from the company. 
Since the puppeteer has a different joint configuration from PAPRAS, PAPRLE controls the follower robot based on end-effector position of the leader device.
Independent of the leader device’s joint configuration, our system allows the operator to control the same follower robot and perform the same task with force feedback.

PAPRLE also accommodates non-puppeteer devices. Figure \ref{fig:leader_joycon} and \ref{fig:leader_dualsense} illustrate the use of gaming controllers as leader for controlling the same PAPRAS arm.
Also, PAPRLE supports VR devices as leader, as shown in Figure \ref{fig:leader_visionpro}, where the operator uses Apple Vision Pro to control two PAPRAS arms.
Furthermore, PAPRLE enables the use of pre-collected datasets for robot control, even when the data is not gathered using PAPRLE itself. Figure \ref{fig:leader_umi} showcases this capability, depicting PAPRAS arms that reproduce motions derived from a dataset collected with the Universal Manipulation Interface (UMI) \cite{umi}, a handheld gripper interface used to record bimanual manipulation demonstrations for training robot policies.
These examples demonstrate the versatility of PAPRLE, enabling seamless control of robots across various leader devices and data sources, facilitating flexible and robust teleoperation for a wide range of tasks and applications.

\subsection{Diverse Follower Configurations}

In this section, we present examples of running diverse configurations of follower robots, shown in Figure \ref{fig:diverse_followers}.

\subsubsection{PAPRAS}
Using PAPRAS, we can configure a variety of limb arrangements. For example, in Figure \ref{fig:follower_table}, two PAPRAS arms are mounted on a table for tabletop tasks. In Figure \ref{fig:follower_kitchen}, the same pair of arms is mounted on a kitchen counter and wall to facilitate loading dishes into a dishwasher from the sink. Figure \ref{fig:follower_orthrus} shows the same arms mounted on a Boston Dynamics Spot robot for mobile manipulation tasks.
These diverse configurations are easily operable by PAPRLE, particularly with the pluggable puppeteer devices. Notably, Figures \ref{fig:follower_table}, \ref{fig:follower_kitchen}, \ref{fig:follower_orthrus} all demonstrate control using the same pair of puppeteer devices, which are simply remounted in different locations to match each setup.

\subsubsection{Humanoid}
PAPRLE can also be applied to the teleoperation of humanoid robots.
Figure \ref{fig:follower_g1_upper} and \ref{fig:follower_g1_full} show the examples of teleoperation of humanoid.
With PAPRLE, users can teleoperate not only the upper body of a humanoid, but also all four limbs simultaneously.

\subsubsection{Commercial Robotic Arm}
PAPRLE further supports the teleoperation of commercial robotic arms beyond PAPRAS.
Figure~\ref{fig:follower_omy} presents an example of teleoperating the Open Manipulator-Y using the PAPRLE framework.
The accompanying video demonstrates all the examples described above.

\begin{figure*}
    \centering
    \begin{overpic}[width=\textwidth]{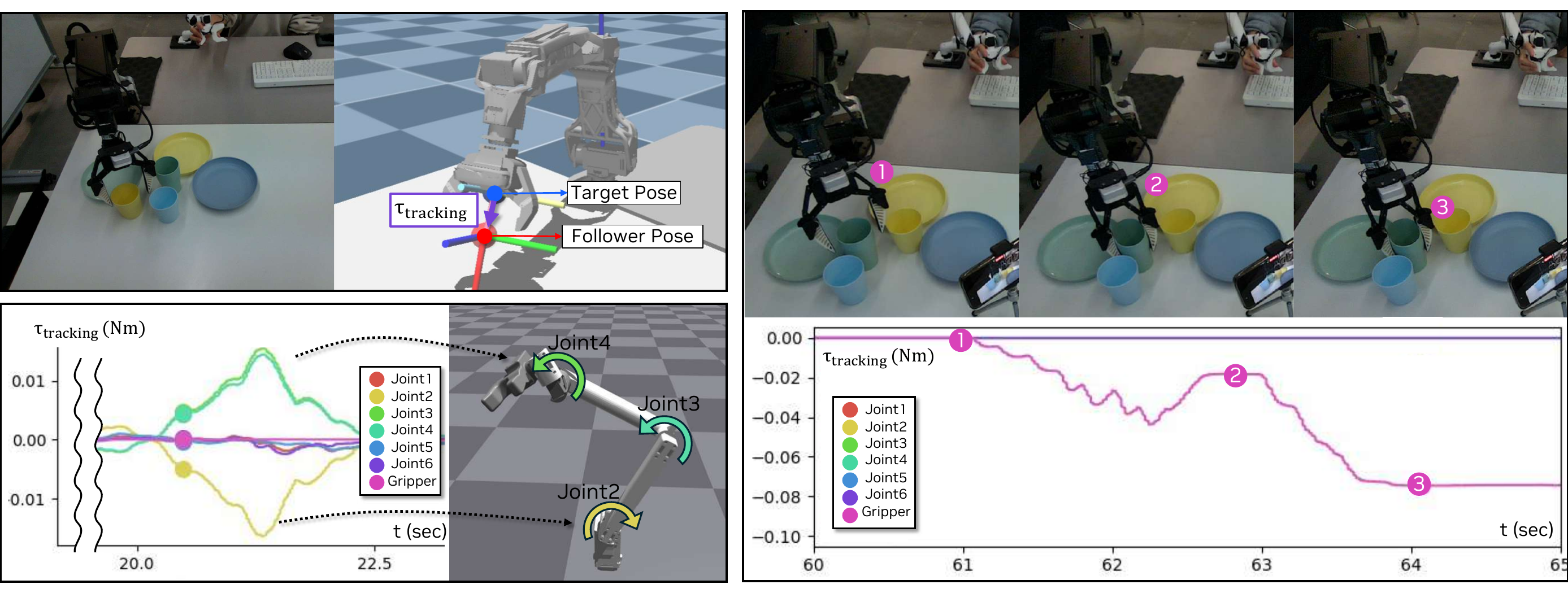}
        \refstepcounter{figure}
        \put(0.2,35.0){\refstepcounter{subfigure}\label{fig:not_following_scn}\colorbox{white!10}{\textbf{[a]}}}
        \put(0.1,1.7){\refstepcounter{subfigure}\label{fig:not_following_grph}\colorbox{white!10}{\textbf{[b]}}}
        \put(47.5,35.1){\refstepcounter{subfigure}\label{fig:gripper_force_feedback}\colorbox{white!10}{\textbf{[c]}}}
    \end{overpic}
    \addtocounter{figure}{-1}
    \caption{\textbf{Examples of the Extrinsic Force Feedback.}}
    \label{fig:force_feedback}
\end{figure*}

\section{Analysis}
\subsection{Runtime}
We analyze the runtime of the proposed PAPRLE system with varying numbers of attached limbs.
Specifically, we measured the average execution time of a single iteration of the for loop  (lines 11–19 in Algorithm~\ref{alg:teleop}) under different command types.
The results are shown in Table \ref{tab:latency}.
When using end-effector (EEF) pose-based commands, the per-step runtime increases, primarily due to the computational overhead of IK and collision checking. 
Since the leader and follower robots have different joint configurations, the system must solve an IK problem to infer appropriate joint positions for the follower, which contributes to the latency. 
Currently, PAPRLE processes each limb sequentially in a for-loop, leading to increased runtime as the number of limbs grows. 
Nonetheless, we observe that even with all four limbs active, PAPRLE maintains a control loop frequency of 50 Hz, which is sufficient for smooth teleoperation and real-time data collection.

\begin{table}[]
\caption{\textbf{Control step duration (ms) of the PAPRLE system for different number of limbs and command types.}}
\label{tab:latency}
\centering
\input{tables/latency}
\end{table}
 
\subsection{Accuracy}
We also analyze the accuracy with which the follower robot tracks the leader’s commands, when they have different kinematic configurations.
We collected several episodes of controlling a 7-DoF PAPRAS robot with two different leaders (UR5 puppeteer, Open Manipulator-Y puppeteer).
The task, illustrated in Figure~\ref{fig:diverse_leaders}, is to sort colored cups into corresponding plates.
By randomizing the initial positions of the cups and plates, we collected 20 episodes for each leader.
Table \ref{tab:accuracy} shows the error between the target end-effector pose command from the leader and the resulting end-effector pose of the follower robot. 
We observed that the majority of errors were under one centimeter, indicating accurate tracking performance despite differences in robot configuration.

\begin{table}[]
\caption{\textbf{Average Positional Error (cm) of Follower Robot} when the puppeteer has different structure.}
\label{tab:accuracy}
\centering
\input{tables/accuracy}
\end{table}

\subsection{Leader feedback}

In this section, we analyze how the force feedback occurs to the leader device, especially when the leader and the follower has different configuration.
In Figure \ref{fig:not_following_scn}, the operator is using an UR5 puppeteer to control PAPRAS.
If the operator moves too fast, because of the velocity limit, the follower robot would not follow the leader command promptly. 
Due to the differences between the target command pose ${\mathbf{T}_t^{\text{follower}}}$ and the follower pose $\mathbf{T}^{\text{follower}}_{t,\text{cmd}}$, the feedback module applies $\mathbf{\tau}_{\text{tracking}}$ to the leader device.
Using Equation \ref{eq:tracking_feedback_delta_q}, the feedback module can calculate joint-level force feedback to leader device, even though the follower does not have the same joint configuration as leaders.
Figure \ref{fig:not_following_grph} shows the plot of how $\mathbf{\tau}_{\text{tracking}}$ appears in each joint of the UR5 puppeteer.
Since the feedback module wants to inform the leader that the follower is down below, each joint of the leader device gets the force feedback into the direction that the end-effector of the leader devices can be lowered.

Figure~\ref{fig:gripper_force_feedback} illustrates how the gripper joint provides force feedback during interaction with objects.
Feedback for the gripper joint is computed at the joint level as a scaled difference between the gripper joint positions of the leader and the follower.
In Figure \ref{fig:gripper_force_feedback}, the force feedback occurs primarily in three stages: first, due to slight speed differences between the leader and follower grippers (stage 1), and second, when the follower gripper makes contact with an object while the operator continues squeezing (stage 2).
At this point, the feedback force becomes noticeably strong, allowing the operator to physically sense resistance and naturally stop squeezing further.
At stage 3, the operator continues to apply force beyond, the position discrepancy grows, leading to stronger feedback.
This gripper feedback can provide the user as the sense of haptic feedback, which can enhance the accurate object manipulation.


\section{Conclusion}
We presented PAPRLE (Plug-And-Play Robotic Limb Environment), a modular and extensible ecosystem for teleoperation that enables flexible pairing between diverse leader devices and follower robots. 
By abstracting control interfaces and supporting both joint-space and task-space modalities, PAPRLE provides a unified framework for intuitive, real-time control across a variety of robot morphologies and environments. 
Its plug-and-play design allows rapid reconfiguration of multi-limb systems, facilitating scalable and reproducible data collection for learning-based robotics. 
Through real-world demonstrations, we validated PAPRLE’s ability to handle diverse teleoperation scenarios with accurate tracking and responsive force feedback, even when leader and follower configurations differ.
As the robotics community continues to pursue embodied AI through large-scale data, we believe PAPRLE offers a foundational platform to accelerate this progress.

\bibliographystyle{IEEEtran}
\bibliography{egbib}



\end{document}

%% file: tables/teleop_pseudo.tex
\begin{algorithmic}[1]
\small{
\Require{\ttfamily{$\text{cfg}_\text{leader}$,$\text{cfg}_\text{follower}$,$\text{cfg}_\text{env}$}}
\State \texttt{leader = init\_leader($\text{cfg}_\text{leader}$)}
\State \texttt{teleop = init\_teleop($\text{cfg}_\text{follower}$)}
\State \texttt{follower = init\_follower($\text{cfg}_\text{follower}$,$\text{cfg}_\text{env}$)}
\State \texttt{feedback = init\_feedback(leader, teleop, follower)}
\State \texttt{feedback.start\_thread()}

\Statex \texttt{\textcolor{lightgray}{\# Initialization Phase}}
\State \texttt{wait\_until\_leader\_starts()}
\State \texttt{cmd = leader.get\_status()}
\State \texttt{qpos = teleop.process(cmd)}
\State \texttt{follower.move\_to(qpos)}

\Statex \texttt{\textcolor{lightgray}{\# Teleoperation Loop}}
\While{\texttt{not shutdown}}
    \State \texttt{cmd = leader.get\_status()}
    \State \texttt{qpos = teleop.process(cmd)}
    \State \texttt{follower.step(qpos)}
    \If{\texttt{leader.requested\_end}}
        \Statex \texttt{\textcolor{lightgray}{\qquad\quad\# Reset}}
        \State \texttt{follower.move\_to(base\_pos)}
        \State \texttt{wait\_until\_leader\_starts()}
        \State \texttt{cmd = leader.get\_status()}
        \State \texttt{qpos = teleop.process(cmd)}
        \State \texttt{follower.move\_to(qpos)}
    \EndIf
\EndWhile
\vspace{-0.25cm}
}
\end{algorithmic}

%% file: tables/supported_devices.tex



\begin{tabular}{@{}cccc@{}}
\toprule
\textbf{Type}             & \textbf{Name} & \textbf{Cmd Type} & \textbf{\# of Limbs}  \\ \midrule
\multirow{4}{*}{Puppeteers} &
  7DOF PAPRAS &
  \multirow{4}{*}{\begin{tabular}[c]{@{}c@{}}Joint Pos, \\ EEF Pose\end{tabular}} &
  \multirow{4}{*}{Unbounded} \\
                          & 6DOF PAPRAS   &                   &                       \\
                          & UR5          &                   &                       \\
                          & OM-Y          &                   &                       \\ \midrule
VR Headset                & VisionPro     & EEF Pose          & $\leq$ 2 \\ \midrule
\multirow{2}{*}{Joystick} & Joycon        & EEF Pose          & 1                     \\
                          & Dualsense     & EEF Pose          & 1         \\ \midrule  
Etc & 
  \begin{tabular}[c]{@{}c@{}}Offline \\ Trajectory\end{tabular} &
  \begin{tabular}[c]{@{}c@{}}Joint Pos, \\ EEF Pose\end{tabular} &
  Unbounded \\ \bottomrule
\end{tabular}

%% file: tables/latency.tex
\begin{tabular}{@{}ccc@{}}
\toprule
\# of Limbs & Direct Joint Mapping & EEF Pose-based \\ \midrule
1           &         1.02          &      3.25          \\
2           &         4.54             &   10.14    \\
4           &        6.27              &        -      \\ \bottomrule
\end{tabular}

%% file: tables/accuracy.tex
\begin{tabular}{@{}ccc@{}}
\toprule
Error (cm)      & UR5 $\rightarrow$ PAPRAS & OMY $\rightarrow$ PAPRAS \\ \midrule
Mean            & 0.47                     & 0.54                     \\
Std             & 0.19                     & 0.19                     \\ \midrule
Max             & 3.04                     & 2.45                     \\
Median          & 0.48                     & 0.57                     \\
99\% Quantile & 0.86                     & 0.87                     \\ 
\bottomrule
\end{tabular}